\documentclass[10pt,twocolumn,letterpaper]{article}

\usepackage{cvpr}
\usepackage{times}
\usepackage{epsfig}
\usepackage{graphicx}
\usepackage{amsmath}
\usepackage{amssymb}

\usepackage[breaklinks=true,bookmarks=false]{hyperref}

\cvprfinalcopy 

\def\cvprPaperID{8370} 

\ifcvprfinal\fi
\begin{document}

\title{Reciprocal Learning Networks for Human Trajectory Prediction}


\author{Hao Sun, Zhiqun Zhao, and Zhihai He\\
University of Missouri\\
{\tt\small \{hshq7,zzhv7,hezhi\}@mail.missouri.edu}
}

\maketitle
\thispagestyle{empty}

\begin{abstract}
We observe that the human trajectory is not only forward predictable, but also backward predictable. Both forward and backward trajectories follow the same social norms and obey the same physical constraints with the only difference in their time directions.
Based on this unique property, we develop a new approach,  called \textit{reciprocal learning}, for human trajectory prediction. 
Two  networks, forward and backward prediction networks, are tightly coupled, satisfying the reciprocal constraint, which allows them to be jointly learned. Based on this constraint, we borrow the concept of adversarial attacks of deep neural networks, which iteratively modifies the input of the network to match the given or forced network output, and develop a new method for network prediction, called \textit{reciprocal attack for matched prediction}. It further improves the prediction accuracy. 
Our experimental results on benchmark datasets demonstrate that our new method outperforms the state-of-the-art methods for human trajectory prediction.
\end{abstract}

\section{Introduction}
\label{sec:intro}

Human motion trajectories and motion patterns are governed by human perception, behavioral reasoning, common sense rules, social conventions, and interactions with others and the surrounding environment. Human can effectively predict short-term body motion of others and respond accordingly. 
The ability for a machine to learn these rules and use them to understand and predict human motion in complex  environments is highly valuable with a wide range of applications in social robots, intelligent systems, and smart environments \cite{luber2010people, mehran2009abnormal}.
The central research question of human trajectory prediction is: \textit{given observed motion trajectories of human, can we predict their future trajectories within a short period of time, for example, 5 seconds, in the future?}

Predicting human motion and modeling their common sense behaviors are a very challenging task \cite{ballan2016knowledge}. An efficient algorithm for human trajectory prediction needs to accomplish the following tasks: (1) \textit{obeying physical constraints of the environment.} To  walk on a feasible terrain and avoid obstacles or other physical constraints, we need to analyze the local and global spatial information surrounding the person and pay attention to important elements in the environment.
(2) \textit{Anticipating movements of other persons or vehicles and their social behaviors.} 
Some trajectories are physically possible but socially unacceptable. Human motions are governed by social norms, such as yielding right-of-way or respecting personal space.  (3) \textit{Finding multiple feasible paths.} There are often  multiple choices of motion trajectories to reach  the destination. This uncertainty poses significant challenges for accurate human trajectory prediction.

\begin{figure}[t]
\begin{center}
 \includegraphics[width=1\linewidth]{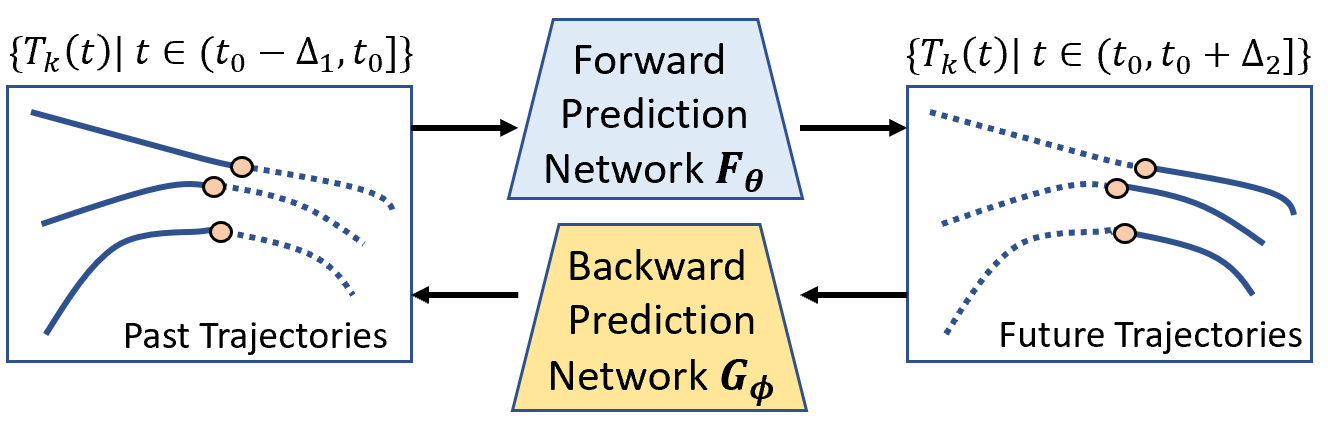}
\end{center}
\caption{Illustration of our idea of reciprocal learning for human trajectory prediction.}
\label{fig:idea}
\end{figure}

Recently, a number of methods based on deep neural networks have been developed for human trajectory prediction \cite{ballan2016knowledge, lee2017desire}. 
Earlier methods have been focused on  learning dynamic patterns of moving agents (human and vehicles)  \cite{ballan2016knowledge} and modeling the semantics of the navigation environment \cite{kitani2012bagnell}.
Methods have been developed to model human-human interactions  \cite{helbing1995social}, understand social acceptability  \cite{bartoli2018context, alahi2016social, lee2017desire}, and model the joint influence of all agents in the scene \cite{gupta2018social}. 
Efforts have also been  taken to predict multiple feasible paths of human \cite{alahi2016social, lee2017desire,xue2018ss}. 

In this work, we propose to explore the unique characteristics of human trajectories and develop a new approach, called \textit{reciprocal learning} for human trajectory prediction.
As illustrated in Figure \ref{fig:idea}, 
we observe that the human trajectory is not only forward predictable, but also backward predictable. Imagine that the time is reversed and the person is traveling backwards.
As discussed in the above, the forward moving trajectories follow the social norm and obey the environmental constraints. So do the backward moving trajectories since the only difference between them is that the time is reversed. 
From the training data, we can train two different prediction networks, the forward prediction network $\mathbf{F}_\theta$ and the backward prediction network $\mathbf{G}_\phi$. These two networks are tightly coupled together, satisfying a reciprocal constraint. For example, using the forward network, we can predict the future trajectory $\mathbf{Y}=\mathbf{F}_\theta(\mathbf{X})$ from the observed or past trajectory $\mathbf{X}$. If the prediction $\mathbf{Y}$ is accurate, then $\mathbf{G}_\phi(\mathbf{Y})$ must be approximately equal to $\mathbf{X}$. 

Based on this  observation and the unique reciprocal constraint, we develop a new approach called \textit{reciprocal network learning} for accurate and robust prediction of human trajectories. We introduce the reciprocal prediction loss and establish an iterative procedure for training these two tightly coupled networks. We borrow the concept of the adversarial attacks of deep neural networks which iteratively modifies the input of the network to match a given target or forced network output. We integrate the reciprocal constraint with the adversarial attack method to develop a new matched prediction method for human trajectory prediction. 
Our experimental results on benchmark datasets demonstrate that our new method outperforms the state-of-the-art methods for human trajectory prediction.

The rest of the paper is organized as follows. Section \ref{sec:rw} reviews related work on human trajectory prediction. The proposed reciprocal network learning and matched prediction are presented in Section \ref{sec:RL}. Section \ref{sec:exp} presents the experimental results, performance comparisons, and ablation studies.
Section \ref{sec:con} summarizes our major contributions and concludes the paper.


\section{Related Work}
\label{sec:rw}
In this section, we review related work, including human-human models and human-scene models for human trajectory prediction, adversarial attacks, and  cycle consistency.

\textbf{(1) Human-human models for trajectory prediction.} 
A number of methods have been developed in the literature to model human social interactions and behaviors in crowded scenes, such as people attempting to avoid  walking into each other.
Helbing and Molnar \cite{helbing1995social} introduced the Social Force Model to characterize social interactions among people in crowded scenes using coupled Langevin equations. In recent methods based on LSTM (Long Short Term Memory) \cite{alahi2016social}, social pooling was introduced to share features and hidden representations between different agents.
The key idea is to merge hidden states of nearby pedestrians to make each trajectory aware of its neighbourhood. 
\cite{bisagno2018group} found out that groups of people moving coherently in one  direction should be excluded from the above pooling mechanism. 
\cite{gupta2018social} used a Generative Adversarial Network (GAN) to discriminate between multiple feasible paths. 
This model is able to capture different movement styles but does not differentiate  between structured and unstructured environments. \cite{vemula2018social} predicted human trajectories using a spatio-temporal graph to model both position evolution and interactions between pedestrians. 

\textbf{(2) Human-scene models for trajectory prediction.} Another  set of methods for human trajectory prediction has focused on  learning the effects of  physical environments. For example, human tend to walk along  the sidewalk, around a tree or other physical obstacles. Sadeghian \textit{et al.} \cite{sadeghian2019sophie}
considered  both  traveled areas and semantic context to predict social and context-aware positions using a GAN (Generative Adversarial Network). \cite{liang2019peeking} extracted multiple visual features, including  each person's body keypoints and the scene semantic map to predict human behavior and model interaction with the surrounding environment. \cite{bartoli2018context} has studied attractions towards static objects, such as artworks, which deflect straight paths in several scenarios such as museums.
\cite{ballan2016knowledge} proposed a Bayesian framework to predict unobserved paths from previously observed motions
and to transfer learned motion patterns to new scenes. In \cite{coscia2018long}, the  dynamics and semantics for long-term trajectory predictions have been studied. Scene-LSTM \cite{manh2018scene} divided the static scene into grids and predicted pedestrian’s location using LSTM. The CAR-Net method \cite{sadeghian2018car} integrated past observations with bird’s eye view images and analyzed them using a two-levels attention mechanism. 

\textbf{(3) Adversarial attacks.} As one of our major contributions, we explore adversarial attacks for network prediction based on reciprocal constraints.  The goal of adversarial attacks is to add small noises on input examples to make them mis-classified by the network. One of the first successful methods to generate adversarial examples is the fast gradient sign method (FGSM) \cite{goodfellow2014explaining}.  Kurakin \textit{et al.} \cite{kurakin2016adversarial} proposed a variant of FGSM called I-FGSM which iteratively applies the FGSM update with a small step size. Note that both FGSM and I-FGSM aim to minimize the Chebyshev distance between the inputs and the generated adversarial examples. Optimization-based methods \cite{szegedy2013intriguing, moosavi2016deepfool, carlini2017towards} have also been developed for  generating adversarial samples.
Our work borrows the idea from FGSM to perform adversarial attacks as a post-processing step on our predicted future trajectories to minimize the self-consistency loss, as explained in  Section \ref{sec:attack}.

\textbf{(4) Cycle consistency learning.} 
Using transitivity as a way to regularize structured data has been studied.  For example, in visual tracking, \cite{kalal2010forward,sundaram2010dense} developed a forward-backward consistency constrain. In language processing, \cite{brislin1970back,he2016dual,twain1971jumping}
studied human and machine translators to verify and improve translations based on back translation and reconciliation mechanisms. Cycle consistency has also been explored in motion analysis \cite{zach2010disambiguating}, 3D shape matching \cite{huang2013consistent}, dense semantic alignment \cite{zhou2016learning,zhou2015flowweb}, depth estimation \cite{godard2017unsupervised,yin2018geonet,zhou2017unsupervised}, and image-to-image translation \cite{bansal2018recycle, zhu2017unpaired}.  CycleGAN \cite{zhu2017unpaired}  introduces a cycle consistence constraint for learning a mapping to translate an image from the source domain into the target domain. In this work, 
we explore the unique characteristics of human trajectories and develop the new approach of reciprocal learning. Our idea is related to the cycle consistency but is quite unique. We introduce the reciprocal loss and design two tightly coupled prediction networks, the forward and backward prediction networks, which are jointly learned based on the reciprocal constraint.

\section{Reciprocal Networks for Human Trajectory Prediction}
\label{sec:RL}
In this section, we present our reciprocal network learning method for human trajectory prediction. 

\subsection{Problem Formulation}
We follow the standard formulation of trajectory forecasting problem in
the literature \cite{vemula2018social, liang2019peeking}. With observed trajectories of all moving agents in the scene, including persons and vehicles, the task is to predict the moving trajectories of all agents for the next period of time in the near future.
Specifically, let $\mathbf{X} = [X_1,X_2, \cdots, X_N]$ be the trajectories of all human in the scene. Our task is to predict the future trajectories of all human 
$\hat{\mathbf{Y}} = [\hat{Y}_1,\hat{Y}_2, \cdots, \hat{Y}_N]$ simultaneously.
The input trajectory of human $n$ is given by
$X_n = (x_n^t, y_n^t)$ for time steps $t=1, 2, \cdots, T_{o}$.
The ground truth of future trajectory is given by 
$Y_n = (x_n^t, y_n^t)$ for time step $t=T_{o}+1, \cdots, T_{p}$.

\subsection{Method Overview}

As illustrated in Figure \ref{fig:idea}, 
in reciprocal learning, we are learning two coupling networks, the forward prediction network $\mathbf{F}_\theta$ which predicts the future trajectories $\mathbf{Y} = \mathbf{F}_\theta(\mathbf{X})$ from the past data $\mathbf{X}$,
and the backward prediction network $\mathbf{G}_\phi$ which predicts the past trajectories $\mathbf{X} = \mathbf{G}_\phi(\mathbf{Y})$ from the future data $\mathbf{Y}$. It should be noted that, during training, both the past and future data are available.
If both networks are well trained, then we should have following two reciprocal consistency constraints:
\begin{eqnarray}
\mathbf{X} &\approx& \mathbf{G}_\phi(\mathbf{F}_\theta(\mathbf{X})), \label{eq-rc1} \\
\mathbf{Y} &\approx& \mathbf{F}_\theta(\mathbf{G}_\phi(\mathbf{Y})). \label{eq:rc-2}
\end{eqnarray}
These two networks are able to help each other to improve the learning and prediction performance.  
Specifically, if the backward prediction network $\mathbf{G}_\phi$ is trained, we can use the reciprocal constraint (\ref{eq-rc1}) to double check the accuracy of the forward prediction network $\mathbf{F}_\theta$ and improve its performance during training. 
Likewise, if the forward prediction network  $\mathbf{F}_\theta$ is trained, we can use (\ref{eq:rc-2}) to improve the training performance of the backward prediction network $\mathbf{G}_\phi$. This results in a tightly coupled iterative learning and performance improvement process between these two prediction networks. Once the forward and backward networks are successfully trained using the reciprocal learning approach,  we develop a new network inference method called \textit{reciprocal attack for matched prediction}. It borrows the concept of  adversarial attacks of deep neural networks 
where the input is iteratively modified such that the network output matches a given target \cite{goodfellow2014explaining}.

Our proposed idea echoes some thoughts in 
CycleGAN \cite{zhu2017unpaired} which presents an approach for learning a  mapping to translate an image from a source domain  to a target domain. They also learn an inverse mapping and introduce the cycle consistence constraint. Our approach is significantly different from this CycleGAN method. We design two tightly coupled prediction networks, the forward and backward prediction networks, which are jointly learned based on the reciprocal constraint. For the testing part, our approach introduces a new reciprocal attack method for matched prediction of human trajectory.

\begin{figure}[htb]
\begin{center}
 \includegraphics[width=0.9\linewidth]{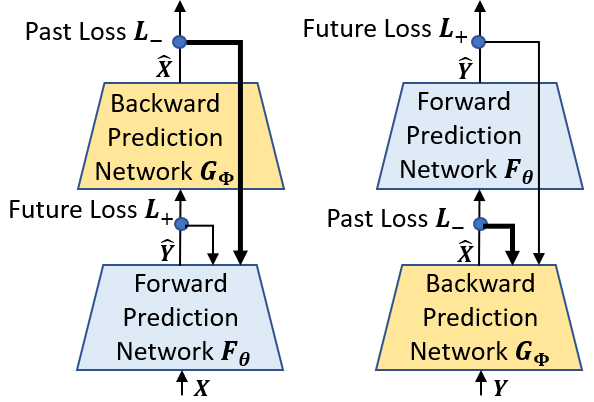}
\end{center}
\caption{The training process of reciprocal learning.}
\label{fig:reciprocal-feedback}
\end{figure}

\subsection{Reciprocal Network Training}

To successfully train the forward and backward prediction networks, we define two loss functions, $J_{-}$ and $J_+$, to measure the prediction accuracy  
of the past and future trajectories. One reasonable choice will be the $L_2$ norm between the original trajectory and its prediction. These two loss functions will be updated alternatively and combined to guide the training of each of these two networks, as illustrated in Figure \ref{fig:reciprocal-feedback}.
For example, when training the forward prediction network $\mathbf{F}_\theta$, the loss function used in existing literature is the prediction error of the future trajectory $L_+$. In reciprocal training, we first pre-train the backward prediction network $\mathbf{G}_\phi$ using the training data with all trajectories reversed in time. We then use this network to map the prediction result of $\mathbf{F}_\theta$,
$\hat{\mathbf{Y}}=\mathbf{F}_\theta(\mathbf{X})$, back to the past trajectory, which is given by 
\begin{equation}
\hat{\mathbf{X}} = \mathbf{G}_\phi(\hat{\mathbf{Y}}) = \mathbf{G}_\phi(\mathbf{F}_\theta(\mathbf{X})). 
\end{equation}
The past trajectory loss is then given by $L_-=||\mathbf{X} - \hat{\mathbf{X}}||_2$.
We refer to this loss as \textit{reciprocal loss}. 
It will be combined with $L_+$ to form the loss function for the forward prediction network $\mathbf{F}_\theta$:
\begin{eqnarray}
\label{eq:J+}
    J_+[\theta] &=& \lambda \cdot L_+ + (1-\lambda)\cdot L_- \nonumber \\
    &=& \lambda\cdot ||\mathbf{Y} - \mathbf{F}_\theta(\mathbf{X})||_2 \\
    &+&   
    (1-\lambda)\cdot ||\mathbf{X} - \mathbf{G}_\phi(\mathbf{F}_\theta(\mathbf{X})) ||_2.
    \nonumber
\end{eqnarray}
Similarly, we can derive the loss function for the backward prediction network $\mathbf{G}_\phi$:
\begin{eqnarray}
\label{eq:J-}
    J_-[\phi] &=& \lambda \cdot L_- + (1-\lambda)\cdot L_+ \nonumber \\
    &=& \lambda\cdot ||\mathbf{X} - \mathbf{G}_\phi(\mathbf{Y})||_2 \\
    &+&   
    (1-\lambda)\cdot ||\mathbf{Y} - \mathbf{F}_\theta(\mathbf{G}_\phi(\mathbf{Y})) ||_2.
    \nonumber
\end{eqnarray}
In reciprocal training, we first pre-train the forward and backward prediction networks independently. Then, these two networks are jointly trained in an iterative manner based on the reciprocal constraint.

\begin{figure}[htb]
\begin{center}
 \includegraphics[width=1\linewidth]{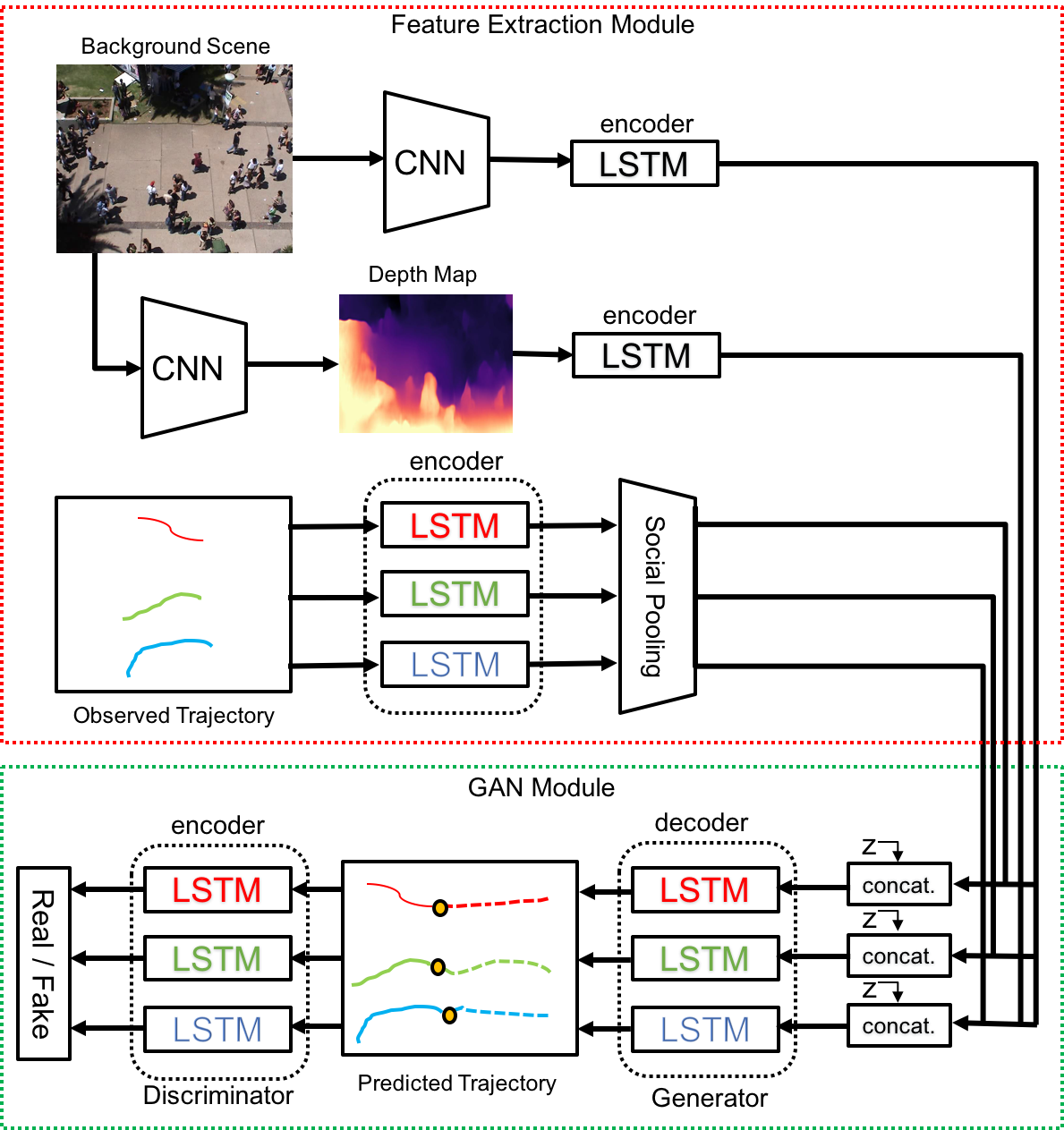}
\end{center}
\caption{Our prediction network has two key components: (1) Feature Extraction Module and (2) LSTM-based GAN module. }
\label{fig:pred_model}
\end{figure}

\subsection{Constructing the Forward and Backward Prediction Networks}

Both the forward and backward networks share the same network structure. In the following, we use the forward prediction network $\mathbf{F}_\theta$ as an example to explain our network design. As illustrated in Figure \ref{fig:pred_model}, we adopt the existing Social-GAN in \cite{gupta2018social} as our baseline prediction network. Our model consists of two key components: (1) a feature extraction module and (2) an LSTM (Long Short Term Memory)-based GAN (Generative Adversarial Network) module.

\subsubsection{Feature Extractor}

Our feature extractor module has three major components to be explained in the following. 
Specifically, we first use the LSTM encoder to capture the temporal pattern and dependency within each trajectory of human $n$ and encode them into a high-dimensional feature $\mathbf{F}_{h}^t(n)$.  
To capture the joint influence of all surrounding human movements on the prediction of the target human $n$, we borrow the idea from \cite{gupta2018social} to build a social pooling module which extracts the joint social feature $\mathbf{F}_{s}^t(n)$ of all  human in the scene to encode the human-human interactions. The relative distance values between the target person and  others  are calculated. These distance vectors are concatenated with the hidden state in the LSTM network for each person and then embedded by an MLP  and followed by a Max-Pooling function to form the joint feature. A maximum number of moving human in the scene is set, whose  default value is  $0$  if the corresponding agent  does not exist at the current time. 

As recognized in \cite{xue2018ss, sadeghian2019sophie}, the environmental context affects the decision of the human in planning its next step of movement. Features of the current scene can be incorporated into the reasoning process. Similar to prior work \cite{sadeghian2019sophie}, we use VGGNet-19 \cite{russakovsky2015imagenet} pre-trained on the ImageNet \cite{russakovsky2015imagenet} to extract the visual feature of background scene $I^t$, which is then fed into an LSTM encoder to compute the hidden state tensor $\mathbf{F}_v^t$.

As a unique feature of our proposed method, we propose to also incorporate the 3D scene depth map into the reasoning process, which also improves the prediction accuracy of human trajectory. This is because the human motion occurs in the original 3D environment. Therefore, its natural behavior and motion patterns are better represented by its 3D trajectory, instead of the 2D image coordinates. For example, the trajectory of a person walking near the camera is much different from that of a person walking far away from the camera due to the camera perspective transform. To address this issue, we propose to infer a depth image from a single image using existing  depth estimation method  \cite{godard2019digging}. We use their pre-trained model to perform monocular depth estimation and obtain the depth map $M_{d}^t$ of scene $I^t$, then use an LSTM to encode it into a depth feature $\mathbf{F}_d^t$. 

\subsubsection{LSTM-based GAN For Trajectory Prediction}

Inspired by previous work \cite{gupta2018social, sadeghian2019sophie}, in this paper we use an LSTM-based Generative Adversarial Network (GAN) module to generate human's future path as illustrated in Figure \ref{fig:pred_model}. The generator is constructed by a decoder LSTM. Similar to the conditional GAN \cite{mehran2009abnormal}, a white noise vector $\mathbf{Z}$ is sampled from a multivariate normal distribution. Then, a merge layer is used in our proposed network which concatenates all  encoded features mentioned above with the noise vector $\mathbf{Z}$. We take this as the input to the LSTM decoder to generate the candidate future paths for each human. The discriminator is built with an LSTM encoder which takes the input as randomly chosen trajectory from either ground truth or predicted trajectories and classifies them as ``real'' or ``fake''. Generally speaking, the discriminator classifies the trajectories which are not accurate as ``fake'' and forces the generator to generate more realistic and feasible trajectories.

Within the framework of our reciprocal learning for human trajectory prediction, let $G^\theta: \mathbf{X} \rightarrow \mathbf{Y}$ and $G^\phi: \mathbf{Y} \rightarrow \mathbf{X}$ be the generators of the forward prediction network $\mathbf{F}_\theta$ and the backward prediction network $\mathbf{G}_\phi$, respectively. $D^\theta$ is 
the discriminator for $\mathbf{F}_\theta$. 
Its input $\mathbf{Y'}$ is randomly selected from either ground truth $\mathbf{Y}$ or the predicted future trajectory $\hat{\mathbf{Y}}$. Similarly, for $D^\phi$
is discriminator for $\mathbf{G}_\phi$.
To train $\mathbf{F}_\theta$ and $\mathbf{G}_\phi$, we combine the adversarial loss with the forward prediction loss  $J_+[\theta]$ and the backward prediction loss $J_-[\phi]$ in Eqs. (\ref{eq:J+}) and (\ref{eq:J-}) together to construct the overall loss function for $\mathbf{F}_\theta$ and $\mathbf{G}_\phi$, respectively:
\begin{equation}
    \mathcal{L}_\theta = L^\theta_{{GAN}} + J_+[\theta], \quad \mathcal{L}_\phi = L^\phi_{{GAN}} + J_-[\phi],
\end{equation}
where adversarial losses $L^\theta_{{GAN}}$ and $L^\phi_{{GAN}}$ are defined as:
\begin{eqnarray}
L^\theta_{{GAN}} &=& \min \limits_{G} \max \limits_{D} \quad
\mathbb{E}_{\mathbf{Y'} \sim p(\mathbf{Y}, \mathbf{\hat{Y}})}[\log D(\mathbf{Y'})] \\
&+&
\mathbb{E}_{\mathbf{X} \sim p(\mathbf{X}) , \mathbf{Z} \sim p(\mathbf{Z})}[\log (1- D(G(\mathbf{X},\mathbf{Z})))],
\nonumber
\end{eqnarray}
\begin{eqnarray}
L^\phi_{{GAN}} &=& \min \limits_{G} \max \limits_{D} \quad
\mathbb{E}_{\mathbf{X'} \sim p(\mathbf{X}, \mathbf{\hat{X}})}[\log D(\mathbf{X'})] \\
&+&
\mathbb{E}_{\mathbf{Y} \sim p(\mathbf{Y}) , \mathbf{Z} \sim p(\mathbf{Z})}[\log (1- D(G(\mathbf{Y},\mathbf{Z})))].
\nonumber
\end{eqnarray}

\subsection{Reciprocal Attack for Matched Prediction of Human Trajectories}
\label{sec:attack}

Once the forward and backward networks are successfully trained with the above loss functions based on the reciprocal learning approach, we are ready to perform prediction of the human trajectories. 
By taking advantage of the reciprocal property of the forward and backward networks, we develop a new network inference method called \textit{reciprocal attack for matched prediction} 
as a post-processing step to further improve the prediction accuracy by making full use of  the current observation.

\begin{figure}[htb]
\begin{center}
 \includegraphics[width=0.9\linewidth]{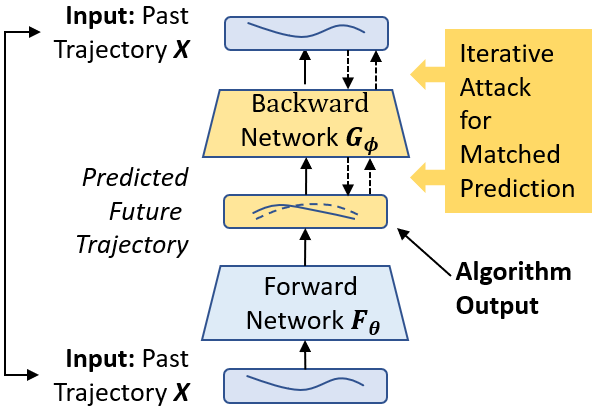}
\end{center}
\caption{Illustration of the proposed attack method.}
\label{fig:reciprocal-attack}
\end{figure}

As illustrated in Figure \ref{fig:reciprocal-attack}, $\mathbf{F}_\theta$ is our trained network for human  trajectory prediction. With the past trajectories $\mathbf{X}$ as input, it predicts the future trajectories 
$\hat{\mathbf{Y}} = \mathbf{F}_\theta(\mathbf{X})$.
During network testing or actual prediction, we do not know the ground truth of the future trajectory. How do we know if this prediction $\hat{\mathbf{Y}}$ is accurate or not? How can we further improve its accuracy? Fortunately, in our reciprocal learning framework, we have another network, the backward prediction network $\mathbf{G}_\phi$, which can be used to map the estimated $\hat{\mathbf{Y}}$ back to the known input $\mathbf{X}$. Our reasoning is that, if 
$\hat{\mathbf{Y}}$ is accurate, then its backward prediction 
$\hat{\mathbf{X}} = \mathbf{G}_\phi(\hat{\mathbf{Y}}) = \mathbf{G}_\phi (\mathbf{F}_\theta(\mathbf{X}))$ should match the original input $\mathbf{X}$.
When the prediction $\hat{\mathbf{Y}}$ is not accurate, we can modify the prediction such that the above matching error is minimized. This leads to the following optimization problem:
\begin{equation}
    \hat{\mathbf{Y}}^* = \arg\min_{\tilde{\mathbf{Y}} =\hat{\mathbf{Y}} + \Delta(t)} 
    ||\mathbf{X} -  \mathbf{G}_\Phi(\tilde{\mathbf{Y}})||_2.
\end{equation}
Here, $\Delta(t)$ is the small perturbation or modification added to the existing prediction result  $\hat{\mathbf{Y}}$. The above optimization procedure aims to find the best modification $\hat{\mathbf{Y}}^* = \hat{\mathbf{Y}} + \Delta(t)$ to minimize the matching error. 

This optimization problem can be solved by adversarial attack methods recently studied in the literature of deep neural network attack and defense.
In this work, we propose to borrow the idea from the  Fast Gradient Sign method (FGSM) developed by Goodfellow \textit{et al.} \cite{goodfellow2014explaining} to perform adversarial attacks. Essentially, it is the same error back propagation procedure as network training. The only difference is that network training modifies the network weights based on error gradients. However, the adversarial attack does not modify the network weights, it propagates the error all the way to the input layer to modify the original input image to minimize the loss. 

This approach uses the sign of the gradient at each pixel to determine its direction of change in its pixel value. In our case, we remove the sign function and directly use the gradient to update the input trajectory.
With the matching error of human trajectories $E=||\mathbf{X} -  \mathbf{G}_\Phi(\tilde{\mathbf{Y}})||_2$, we can perform multiple iterations of the modified FGSM attack on the prediction $\hat{\mathbf{Y}}$ such that the matching error is minimized. 
At iteration $m$, the attacked trajectory (input) is given by 
\begin{equation}
    \hat{\mathbf{Y}}^m = \hat{\mathbf{Y}}^{m-1} + \ 
    \epsilon\cdot \nabla_{\hat{\mathbf{Y}}} {E}(\mathbf{X},\hat{\mathbf{Y}}^{m-1}),
\end{equation}
with $\hat{\mathbf{Y}}^0 = \hat{\mathbf{Y}}$. 
$\epsilon$ is the magnitude of attacks \cite{goodfellow2014explaining}. 
Intuitively, the updated trajectory $\hat{\mathbf{Y}}^m$ will minimize ${E}$. 
We then perform an exponential average of $\{\hat{\mathbf{Y}}^m\}$ to obtain the improved prediction
\begin{equation}
    \hat{\mathbf{Y}}^* = \left[\sum_{m=1}^M e^{\alpha \cdot m} \cdot\hat{\mathbf{Y}}^m\right] / \sum_{m=1}^M e^{\alpha \cdot m},
\end{equation}
where $M$ is the total iterations and $\alpha$ is a constant to control the relative weights between these different iterations of attacks. 
Its value is chosen based on heuristic studies. In our experiments, we set $\alpha=0.1$.

\section{Experimental Results}
\label{sec:exp}
In this section, we present our experimental results, performance comparison with state-of-the-art methods, and ablation studies.

\subsection{Benchmark Datasets}
The comparison and ablation experiments are performed on ETH \cite{pellegrini2010improving} and UCY \cite{leal2014learning} datasets which contain real world human trajectories and various natural human-human interaction circumstance. In total, 5 sub-datasets, ETH, HOTEL, UNIV, ZARA1 and ZARA2, 
are included in these two datasets. Each set contains bird-view images and 2D locations of each person. In total there are 1536 persons in these five sets of data. They contain challenging situations, including human collision avoidance, human crossing each other and group behaviors \cite{sadeghian2019sophie}. 

\subsection{Implementation Details}
Our GAN model is constructed using the LSTM for the encoder and decoder. The generator and discriminator are trained iteratively with the Adam optimizer.  We choose the batch size of $64$ and the initial learning rate of $0.001$. The whole model is trained for $200$ epochs.
The trajectories are embedded using a single layer MLP with dimension of $16$. 
The encoder and decoder for the generator use an LSTM with the hidden state's dimension of $32$. In the LSTM encoder for the discriminator, the hidden state's dimension is $48$.  The maximum number of human surrounded with the target human is set as $32$. This value is chosen since in all datasets, none of them has more than $32$ human in any frame.  For the depth map extraction, we use the pre-trained model ``monodepth2'' from \cite{godard2019digging} and the depth feature is embedded using a single layer MLP with an embedding dimension of $16$. The weight for our loss function is $\lambda = 0.5$. We perform the reciprocal attack for $20$ iterations, the perturbation $\epsilon$ is set as $-0.05$.

\subsection{Evaluation Metrics and Methods}

We use the same error metrics in \cite{alahi2016social, pellegrini2009you} for performance evaluations. 
(1) Average Displacement Error (ADE) is the average $L_2$ distance between the ground truth and our prediction over all predicted time steps from $T_o+1$ to $T_p$. (2) Final Displacement Error (FDE) is the Euclidean distance between the predicted final destination and the true final destination at end of the prediction period $T_{p}$. They are defined as:
    \begin{equation}
        \text{ADE} = \frac{\sum\limits_{i\in \Psi} \sum\limits^{T_{p}}_{t=T_{o} +1} \sqrt{((\hat{x}_t^i,\hat{y}_t^i)-(x_t^i,y_t^i))^2}} {\left|\Psi  \right| \cdot T_{p}},
    \end{equation}
    \begin{equation}
    \text{FDE} = \frac{\sum\limits_{i\in \Psi} \sqrt{((\hat{x}_{T_{p}}^i, \hat{y}_{T_{p}}^i)-(x_{T_{p}}^i, y_{T_{p}}^i))^2}}{\left|\Psi \right|},
    \end{equation}
where $(\hat{x}_t^i,\hat{y}_t^i)$ and $(x_t^i,y_t^i)$ are the predicted and ground truth coordinates for human $i$ at time $t$, $\Psi$ is the set of human and $\left|\Psi \right|$ is the total number of human in the test set.

Following previous papers \cite{alahi2016social, gupta2018social, sadeghian2019sophie}, we use the similar leave-one-out evaluation methodology. Four datasets are used for training and the remaining one is used for testing. Given the human trajectory for the past 8 time steps (3.2 seconds), our model predicts the future trajectory for next 12 time steps (4.8 seconds).   

	\begin{table*}[htb]
		\caption{Comparisons of different methods on ETH (Column 2 and 3) and UCY (Column 4-6) datasets on the task of predicting 12 future time steps, given the previous 8 time steps. Error metrics reported are ADE / FDE in meter scale.}
		\begin{center}
		\resizebox{1.5\columnwidth}{!}{
			\begin{tabular}{l c c c c c c}
				\hline
				Method & ETH & HOTEL & UNIV & ZARA1 & ZARA2 & Avg \\ 
				\hline
			    Linear & 1.33 / 2.94 & 0.39 / 0.72 & 0.82 / 1.59 & 0.62 / 1.21 & 0.77 / 1.48 & 0.79 / 1.59  \\
				LSTM  &1.09 / 2.14 & 0.86 / 1.91 &0.61 / 1.31 & 0.41 / 0.88 & 0.52 / 1.11 & 0.70 / 1.52  \\
				S-LSTM\cite{alahi2016social} & 1.09 / 2.35 & 0.79 / 1.76 & 0.67 / 1.40 & 0.47 / 1.00 & 0.56 / 1.17  & 0.72 / 1.54    \\
				S-GAN\cite{gupta2018social}  & 0.81 / 1.52 & 0.72 /1.61 & 0.60 / 1.26 & 0.34/ 0.69 & 0.42 / 0.84 & 0.58 / 1.18 \\
				S-GAN-P\cite{gupta2018social}  & 0.87 / 1.62 & 0.67 / 1.37 & 0.76 / 1.52 & 0.35 / 0.68 & 0.42 / 0.84 & 0.61 / 1.21 \\
				SoPhie\cite{sadeghian2019sophie} & 0.70 / 1.43 &0.76 / 1.67 & 0.54 / 1.24& 0.30 / 0.63 & 0.38 / 0.78  & 0.54 / 1.15  \\
				Next\cite{liang2019peeking} & 0.73 / 1.65 & \textbf{0.30} / \textbf{0.59} & 0.60 / 1.27 & 0.38 / 0.81 & 0.31 /  0.68  & 0.46 / 1.00  \\
				\hline
				Ours & \textbf{0.69} / \textbf{1.24} & 0.43 / 0.87 & \textbf{0.53} / \textbf{1.17} & \textbf{0.28} / \textbf{0.61}  & \textbf{0.28} / \textbf{0.59}  & \textbf{0.44} / \textbf{0.90}\\
				\hline	
			\end{tabular}}
		\end{center}
		\label{tab:eth-ucy}
	\end{table*}

\subsection{Comparison with Existing Methods}
We compare our method with the following state-of-the-art methods:
(1) \textit{Linear}: This method applies a linear regression to estimate linear parameters by minimizing the least square error \cite{gupta2018social}.
(2) \textit{LSTM}: This is the baseline model for the LSTM method, which does not consider any human-human interaction or background scene information.
(3) \textit{S-LSTM} \cite{alahi2016social}: This method models each human by an LSTM and proposes a social pooling mechanism with the hidden states of human within a certain grid at each time step. 
(4) \textit{S-GAN} \cite{gupta2018social}: This is one of the first GAN-based methods. During the pooling stage, all human  in the scene are considered. \textit{S-GAN} and \textit{S-GAN-P} are different only in whether the pooling mechanism is applied or not. The method chooses the best trajectory from 20 network predictions as the final test result.
(5) \textit{SoPhie} \cite{sadeghian2019sophie}: This work implements the so-called physical constrain described by  background scene features. Also the attention mechanism is introduced in this GAN-based method. 
(6) \textit{Next} \cite{liang2019peeking}: This method implements a multiple feature pooling LSTM-based predictor. In the test part, besides using a single model, the paper follows  \cite{gupta2018social} to train 20 different models using random initialization. They reported both ``single model'' and ``20 outputs'' evaluation results in the paper. In our comparison, we  select the best results from these two parts.

\subsection{Quantitative Results}
Table \ref{tab:eth-ucy} shows the comparison results of our method against existing methods  on  performance metrics ADE and FDE in meter scale. 
We follow the prior work \cite{gupta2018social} to choose the best prediction among multiple samples in $L_2$ norm for quantitative evaluation. We can see that our method outperforms all other methods except the Hotel dataset against the \textit{Next} method. 
The \textit{Linear} model generally performs the worst. It can only predict the straight trajectory and suffers from degraded performance in complicated human-human and human-environment interactions. The \textit{LSTM} approach performs better than \textit{Linear} method since it can handle more complicated trajectories. \textit{S-LSTM} also outperforms the \textit{Linear} model since it uses the social pooling mechanism, but it performs worse than \textit{LSTM}. According to \cite{gupta2018social}, the \textit{S-LSTM} \cite{alahi2016social} is trained on a synthetic dataset and fine-tuned on the real dataset to improve the accuracy. 

To evaluate the performance of our method in predicting feasible paths in crowded scenes, we follow the procedure in previous papers \cite{sadeghian2019sophie} to report a new evaluation metric which is the percentage of \textit{near-collisions} among humans. A collision is defined  when the euclidean distance between two human is smaller than 0.1m.
We compute the average percentage of human near-collision in each frame of ETH and UCY datasets.
The comparison results against the \textit{Linear} , \textit{S-GAN} and \textit{SoPhie} methods are shown in Table \ref{tab:collision}. 
We can see that our method outperforms these three methods on the ETH, HOTEL, and ZARA2 datasets, producing less human collisions in the future time. For the other two datasets, UNIV and ZARA1, \textit{S-GAN} and \textit{SoPhie} are slightly better than ours. However, they suffer from significant performance degradation on other datasets. 
Overall, the experimental results demonstrate that our method can predict better physical and socially acceptable paths when compared to these existing methods.

\begin{table}[htb]
\caption{Average percentage of colliding human for each scene in ETH and UCY datasets. A human collision is defined and detected as the Euclidean distance between two human is less than 0.1m \cite{sadeghian2019sophie}. The first column represents the ground truth.}
\begin{center}
\resizebox{0.9\linewidth}{!}{
\begin{tabular}{l |c |c |c |c |c}
\hline
 & GT & Linear & S-GAN & SoPhie  & Ours \\
\hline
ETH  & 0.000 & 3.137 & 2.509 & 1.757 & \textbf{1.512}  \\ 
HOTEL & 0.092 & 1.568 & 1.752 & 1.936 & \textbf{1.547}  \\
UNIV  & 0.124 & 1.242 & \textbf{0.559} & 0.621 & 0.563  \\
ZARA1 & 0.000 & 3.776 & 1.749 & \textbf{1.027} & 1.094  \\
ZARA2 & 0.732 & 3.631 & 2.020 & 1.464 & \textbf{1.252}  \\ \hline
Avg &0.189 & 2.670 & 1.717 & 1.361 & \textbf{1.194}  \\
\hline
\end{tabular}}
\end{center}
\label{tab:collision}
\end{table}

\begin{table*}[htb]
\caption{Ablation experiments of our full algorithm without different components. Error metrics reported are ADE / FDE in meter scale.}
\begin{center}
\resizebox{1.7\columnwidth}{!}{
\begin{tabular}{l| c| c| c| c| c}
\hline
Method & ETH & HOTEL & UNIV & ZARA1 & ZARA2\\
\hline
Our Method (Full Algorithm) & 0.69 / 1.24 & 0.43 / 0.87 & 0.53 / 1.17 & 0.28 / 0.61  & 0.28 / 0.59\\ \hline
\quad\quad - Without Reciprocal Learning &0.73 / 1.31 &0.49 / 0.97 &0.60 / 1.22 & 0.38 / 0.73 & 0.36 / 0.70 \\
\quad\quad - Without Depth Features &0.71 / 1.30 & 0.43 / 0.88 &0.56 / 1.19 & 0.31 / 0.63 & 0.31 / 0.62 \\
\quad\quad - Without Reciprocal Attacks &0.70 / 1.26 &0.45 / 0.90 &0.55 / 1.18 & 0.32 / 0.65 & 0.30 / 0.61\\
\hline
\end{tabular}
}
\end{center}
\label{tab:ablation}
\end{table*}

\subsection{Ablation Studies}
To systematically evaluate our method and study the contribution of each algorithm component, we perform a number of ablation experiments. 
Our algorithm has three major new components, the reciprocal learning, the incorporation of 3D depth map features, and the reciprocal attacks for matched prediction. 
In the first row of Table \ref{tab:ablation}, we list the ADE and FDE results for our method (full algorithm). The second row shows the results for our method without reciprocal training. The third row shows results without depth map features. The last row shows results without reciprocal attacks for prediction. 
We can clearly see that each algorithm component is contributing to the overall performance. 

With the reciprocal consistency constraint, during training, our model forces the backward predicted trajectory to be consistent with the observed past trajectory, thus the predicted future trajectory which is the input of the backward network will be forced to be closer to the ground truth. 
Results show the benefit of the depth feature since it can help the model to better understand human behavior and the background scene context.
The reciprocal attack mechanism modifies the predicted trajectory in an iterative manner to match the original trajectory with the backward prediction network. 

\subsection{Qualitative Results}
\begin{figure}[htb]
\begin{center}
\includegraphics[width=1\linewidth]{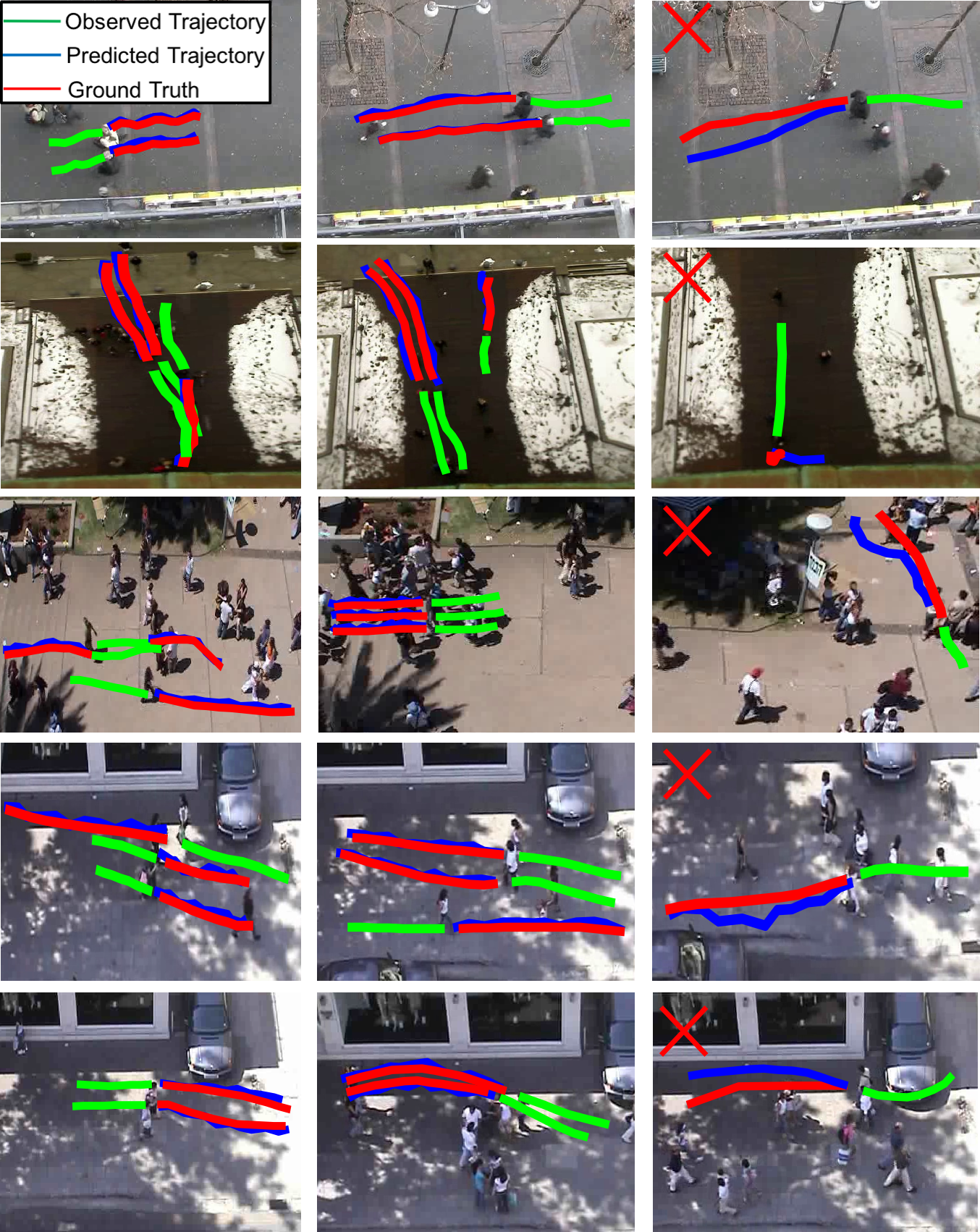}
\end{center}
  \caption{Illustration of our method predicting trajectories of future 12 time steps, given the observed trajectories of past 8 time steps. The results for ETH, HOTEL, UNIV and ZARA1 and ZARA2 are shown in rows 1 to 5, respectively. We show examples where our model successfully predicts the trajectories with small errors  in first two columns. The last column shows some failure cases. Note that, we cropped and resized the original image for better visualization.}
\label{fig:result}
\vspace{-5mm}
\end{figure}

Figure \ref{fig:result} shows successful and failure examples of our predicted trajectories from ETH, HOTEL, UNIV, ZARA1 and ZARA2 datasets in each row.
Following prior work S-GAN \cite{gupta2018social}, we show the best predicted trajectory among 20 model outputs in the figure. 
The first two columns show scenarios that our proposed method is able to correctly predict the future path. According to the background scene, we can see that our method can ensure that each human path follows the physical constrains of the scene, such as walking around obstacles, \textit{e.g.} trees, and staying on sidewalks. Our method also shows the decent prediction results under human-human interactions circumstance. When persons walk in a crowded road, they can avoid each other when they merge from various directions and then walk towards a common direction.

The last column in Figure \ref{fig:result} shows some failure cases which have relatively large error rates. For example, we see human slowing down or even stops for a while, or human taking a straight path rather than making a detour around the obstacles.
Nevertheless, in most case, our method still can predict the plausible path, even though the predicted path is not quite same as the ground truth. For example, for the first, third and fifth cases in the last column, in our prediction paths, the target human are trying to walk around another human or the tree in the road, which are quite reasonable in practice. 

\section{Conclusion and Major Contributions}
\label{sec:con}
In this paper, we have explored the unique characteristics of human trajectories and developed a new approach, reciprocal network learning, for human trajectory prediction.
Extensive experimental results have demonstrated our approach achieves the state-of-art performance on  public  benchmark datasets.

The \textbf{major contributions} of this work can be summarized as follows.
(1) We have established a forward and backward prediction network structure for human trajectory prediction, which satisfies the reciprocal prediction constraints. 
(2) Based on this constraint, we have developed a reciprocal learning approach to jointly train these two prediction networks in an collaborative and iterative manner. 
(3) Once the network is successfully trained, we have developed a new approach for network inference by integrating the concept of adversarial attacks with the reciprocal constraint. It is able to iteratively refine the predicted trajectory by the forward network such that the reciprocal constraint is satisfied.  
(4) Our ablation studies have shown that the proposed new approach is very effective with significant contributions to the overall performance of our method, which outperforms other state-of-the-art methods in the literature.

\section*{Acknowledgement}
This work was supported in part by National Science Foundation under grants 1647213 and 1646065.
Any opinions, findings, and conclusions or recommendations expressed in this material are those of the authors and do not necessarily reflect the views of the National Science Foundation.

{\small
\bibliographystyle{ieee_fullname}
\bibliography{mybib}
}

\end{document}